\documentclass{article}

\usepackage{microtype}
\usepackage{graphicx}
\usepackage{subfigure}
\usepackage{booktabs} 
\usepackage[table,x11names,dvipsnames,table]{xcolor}
\usepackage{multirow}
\usepackage{hyperref}

\usepackage[noend]{algpseudocode}

\usepackage[accepted]{icml2024}

\usepackage{amsmath}
\usepackage{amssymb}
\usepackage{mathtools}
\usepackage{amsthm}

\usepackage[capitalize,noabbrev]{cleveref}

\theoremstyle{plain}

\theoremstyle{definition}

\theoremstyle{remark}

\usepackage[textsize=tiny]{todonotes}

\icmltitlerunning{Robust Optimization in Protein Fitness Landscapes Using RL in Latent Space}

\begin{document}

\twocolumn[
\icmltitle{Robust Optimization in Protein Fitness Landscapes Using \\ Reinforcement Learning in Latent Space}



\icmlsetsymbol{equal}{*}

\begin{icmlauthorlist}
\icmlauthor{Minji Lee}{equal,kaistcs}
\icmlauthor{Luiz Felipe Vecchietti}{equal,ds}
\icmlauthor{Hyunkyu Jung}{kaistcs,ds}
\icmlauthor{Hyunjoo Ro}{bcs}
\icmlauthor{Meeyoung Cha}{kaistcs,ds}
\icmlauthor{Ho Min Kim}{kaistbio,bcs}
\end{icmlauthorlist}

\icmlaffiliation{kaistcs}{School of Computing, KAIST, Daejeon, South Korea}
\icmlaffiliation{ds}{Center for Mathematical and Computational Sciences, Institute for Basic Science, Daejeon, South Korea}
\icmlaffiliation{kaistbio}{Department of Biological Sciences, KAIST, Daejeon, South Korea}
\icmlaffiliation{bcs}{Center for Biomolecular and Cellular Structure, Institute for Basic Science, Daejeon, South Korea}

\icmlcorrespondingauthor{Meeyoung Cha}{meeyoungcha@kaist.ac.kr}
\icmlcorrespondingauthor{Ho Min Kim}{hm\_kim@kaist.ac.kr}

\icmlkeywords{Machine Learning, ICML}

\vskip 0.3in
]



\printAffiliationsAndNotice{\icmlEqualContribution} 

\begin{abstract}
Proteins are complex molecules responsible for different functions in nature. Enhancing the functionality of proteins and cellular fitness can significantly impact various industries. However, protein optimization using computational methods remains challenging, especially when starting from low-fitness sequences. We propose LatProtRL, an optimization method to efficiently traverse a latent space learned by an encoder-decoder leveraging a large protein language model. To escape local optima, our optimization is modeled as a Markov decision process using reinforcement learning acting directly in latent space. We evaluate our approach on two important fitness optimization tasks, demonstrating its ability to achieve comparable or superior fitness over baseline methods. Our findings and in vitro evaluation show that the generated sequences can reach high-fitness regions, suggesting a substantial potential of LatProtRL in lab-in-the-loop scenarios.
\end{abstract}
\section{Introduction}

Proteins mediate the fundamental processes of life. Improving the proteins' functions or cellular fitness is crucial for industrial, research, and therapeutic applications \citep{yang2019machine, huang2016coming}. 
One powerful approach to this is \textit{directed evolution}, the iterative process of performing random mutations and screening proteins with desired phenotypes. However, the protein sequence space of possible combinations of 20 amino acids is too vast to search exhaustively in the laboratory, even with high-throughput screening from diversified libraries \citep{huang2016coming}. Therefore, computational methods have been proposed to improve the success rate of fitness optimization by generating optimized sequences or predicting beneficial mutations \citep{brookes2019conditioning, sinai2020adalead, kirjner2023optimizing}. 
When experimental data is available, fitness predictors have been trained to act as surrogate models for optimization \citep{rao2019evaluating, dallago2021flip}.

In our work, we consider protein fitness optimization in an \textit{active learning} setting (Algorithm~\ref{alg:activelearning}) starting from low-fitness sequences. The model iteratively proposes optimized sequences, gets feedback from a black-box oracle, and updates its belief on the fitness landscape, \textit{i.e.}, mapping between sequence and fitness, for the next optimization round. For example, Bayesian optimization (BO) methods \cite{brookes2019conditioning, belanger2019biological} update their acquisition function after each round. In this paper, we propose an optimization of sequences via reinforcement learning (RL) directly in a latent representation space rather than in the protein sequence space. Our protein fitness optimization framework (Figure~\ref{fig:overview}) works as follows:

\begin{itemize}
    \item \textbf{State}. We introduce a variant encoder-decoder (VED), which reduces a protein sequence into a low-dimensional representation using a pre-trained protein language model (pLM). In the decoder, we introduce a prompt-tuning approach to recover the sequence from predicted embeddings. By treating the learned representation as \textit{states}, we can use the knowledge of a large pre-trained language model and detach the representation learning from optimization. 
    \item \textbf{Action}. 
    Generating sequences via single mutations can lead to challenges in exploration, particularly for proteins where multiple mutations are required to acquire high fitness.
    Therefore, we define actions as perturbations in a latent space. The optimized representation generated by the policy is decoded back to the sequence space using the variant encoder-decoder.
    \item \textbf{Optimization}. The protein fitness landscape is often \textit{rugged} \cite{szendro2013quantitative}, exhibiting multiple peaks surrounded by low fitness valleys. When starting from a low fitness sequence, it is important to cross the valleys and escape local optima. We use a Markov decision process to model fitness optimization, where we design the protein over several timesteps and train the RL policy to maximize the expected future rewards. 
\end{itemize}

In the optimization process, at each timestep, the policy updates the latent representation by making small perturbations to maximize fitness, \emph{i.e., walking} uphill through the local landscape until the end of an episode. We also propose three essential components to efficiently model this RL setting. First, we store the previously found maxima in a \textit{frontier buffer} and sample initial states from it. Second, we give negative feedback to the RL policy based on the number of mutations per step as \textit{calibrating steps}. Third, we add a \textit{constrained decoding} strategy by only applying the most probable predicted mutations. 

We evaluate our method, named \textbf{LatProtRL}, in two fitness optimization tasks with accurate in silico oracles available: optimizing the green fluorescent protein (GFP) and the functional segment of adeno-associated virus (AAV). Our results show that LatProtRL design sequences with comparable or higher fitness than previous methods. We showcase that our designs successfully escape local optima and reach high-fitness regions in the experimental data for GFP while other methods fail. We also provide ablation studies on the state/action modeling, the proposed components, and a single-round optimization setting where we use a fitness predictor as a surrogate model for the black-box oracle. Code is available in \href{https://github.com/haewonc/LatProtRL}{https://github.com/haewonc/LatProtRL}.
\section{Related Works}
\label{sec:related}

\textbf{Protein Representation Learning} aims to learn compact and expressive features describing the protein. Since a protein can be represented as a sequence of amino acids ($N=20$), language models such as BERT \citep{devlin2018bert} are widely used~\citep{brandes2022proteinbert, lin2022language}. \citet{rives2021biological} introduced ESM-2, a pLM trained with 250 million sequences, which produces representations expressing biological properties and reflecting protein structure. Similarly to our work, previous literature explored learning latent representations for optimization~\citep{gomez2018automatic, stanton2022accelerating}.

\textbf{Protein Fitness Prediction}. 
The pLM representation can be generalized across different applications, achieving state-of-the-art results for zero-shot \citep{notin2022tranception} and supervised \citep{rao2019evaluating} fitness prediction. \citet{notin2022tranception} proposed an autoregressive transformer architecture for fitness prediction that achieved high performance in 87 protein datasets. Traditional machine learning methods and CNN-based architectures have been applied to fitness prediction as in~\citet{yang2019machine}.

\begin{algorithm}[t!]
\caption{Fitness optimization as active learning}
\label{alg:activelearning}
    \begin{algorithmic}[1]
    \State Set of measured sequences $\mathcal{S}$
        \For{$E$ rounds}
            \State Propose $N_\text{oracle\_calls}$ sequences
            \State Evaluate the proposed sequences using an oracle $q$
            \State Augment $\mathcal{S}$ with evaluated sequences
            \State Update knowledge on fitness landscape using $\mathcal{S}$
        \EndFor
    \end{algorithmic}
\end{algorithm}

\subsection{Protein Fitness Optimization} 
\textbf{Reinforcement Learning}. \citet{angermueller2019model} proposed a variant of proximal policy optimization~\citep{schulman2017proximal} that trains an offline policy using previously measured sequences to autoregressively generate the optimized sequence. \citet{wang2023self} explored self-play and Monte Carlo tree search for protein and peptide design.

\textbf{Bayesian Optimization}. BO is extensively studied for fitness optimization in \citet{romero2013navigating, belanger2019biological, terayama2021black, swersky2020amortized}. Among these approaches, \citet{stanton2022accelerating} optimize the sequence with respect to multiple objectives directly in a latent space by training a denoising autoencoder that learns representations for corrupted sequences. Compared to \citet{stanton2022accelerating}, our approach does not use the corrupt-and-denoise idea and defines optimization as an episodic task to use reinforcement learning. 

\textbf{Energy-Based Models (EBM)}. \citet{kirjner2023optimizing} proposed a method to smooth the fitness landscape and sample sequences based on the gradients of a differentiable fitness predictor trained on experimental data. They show state-of-the-art results in fitness optimization when a differentiable predictor is available. As an ablation study, we also evaluate the proposed methodology in this setting. \citet{frey2023protein} proposed a discrete EBM framework for antibody design that learns a smoothed energy function and samples from this data manifold using Langevin Markov chain Monte Carlo (MCMC). A denoiser network is also trained to recover denoised sequences from noisy inputs. Compared to \citet{frey2023protein}, our decoder recovers sequences from a latent space, and not from noisy amino acid distributions.

\textbf{Evolutionary Algorithms}. \citet{sinai2020adalead} proposed a rollout method that greedily mutates the sequences using a predictor trained based on the oracle feedback. \citet{ren2022proximal} proposed an exploration algorithm named PEX that prioritizes a lower number of mutations between the optimized sequence and the wild-type sequence.

\textbf{Generative Models} are explored in \citet{schmitt2022prediction,jain2022biological, kim2023bootstrapped} to search and sample optimized sequences. \citet{jain2022biological} used GFlowNets with the focus on the generation of diverse candidates.

\section{Methodology}

\subsection{Problem Formulation}

We define a protein $\boldsymbol{x} = (x_1, \cdots, x_L)$ as a sequence of amino acids with length $L$, where $x_i \in \mathcal{V}$ is the amino acid at the $i$-th position, and $\mathcal{V}$ is the vocabulary of 20 amino acids. In our in silico optimization tasks, we assume the availability of a large-scale dataset of variant sequences with fitness measurements $\mathcal{D^*}=\{(\boldsymbol{x}^{1}, f(\boldsymbol{x}^{1})), \cdots, (\boldsymbol{x}^{N},f(\boldsymbol{x}^{N}))\}$ to train an in silico oracle $q_\theta$, where $N$ is the number of sequences in the dataset and $f(\boldsymbol{x})$ represents the fitness measurement for sequence $\boldsymbol{x}$. Here, we use the term \textit{fitness} to represent a desired protein functionality.
We consider a set of low-fitness proteins $\mathcal{D}$ sampled from $\mathcal{D}^*$ to only contain proteins with fitness value lower than a certain percentile of $\mathcal{D}^*$. $\mathcal{D}$ is used for optimization by the proposed method (i) to set initial states; (ii) to train the encoder-decoder; and (iii) to train a fitness predictor $g_\phi$ if applicable. The main objective is to generate $M$ sequences with high fitness, and desirably, high diversity. Algorithm~\ref{alg:activelearning} describes the general procedure of  active learning. This setting is readily applicable to in vitro oracles when only a low-fitness dataset $\mathcal{D}$ is available.

\subsection{Optimization in Latent Space}  

We train a pair of encoder $\mathcal{E}_\theta$ that maps the protein sequence to a latent space with reduced dimensionality $z=\mathcal{E}_\theta(\boldsymbol{x})$, and decoder $\mathcal{R}_\phi$ that decodes this representation back to sequence space. We optimize the protein by performing a perturbation in the latent representation $z$ of the current sequence $\boldsymbol{x}$ and decoding the perturbed representation $z'$ back to sequence space $\boldsymbol{x}' = \mathcal{R}_\phi (z')$. Using a latent space formulation can efficiently detach the representation learning step in optimization and allow the use of pre-trained embedding models. It also allows flexibility as a perturbation in latent space can lead to multiple mutations at each optimization step, compared to generating \citep{angermueller2019model} or mutating a single position per step \citep{belanger2019biological, kirjner2023optimizing}. 

\begin{figure}[t!]
    \centering
  \includegraphics[width=.9\columnwidth]{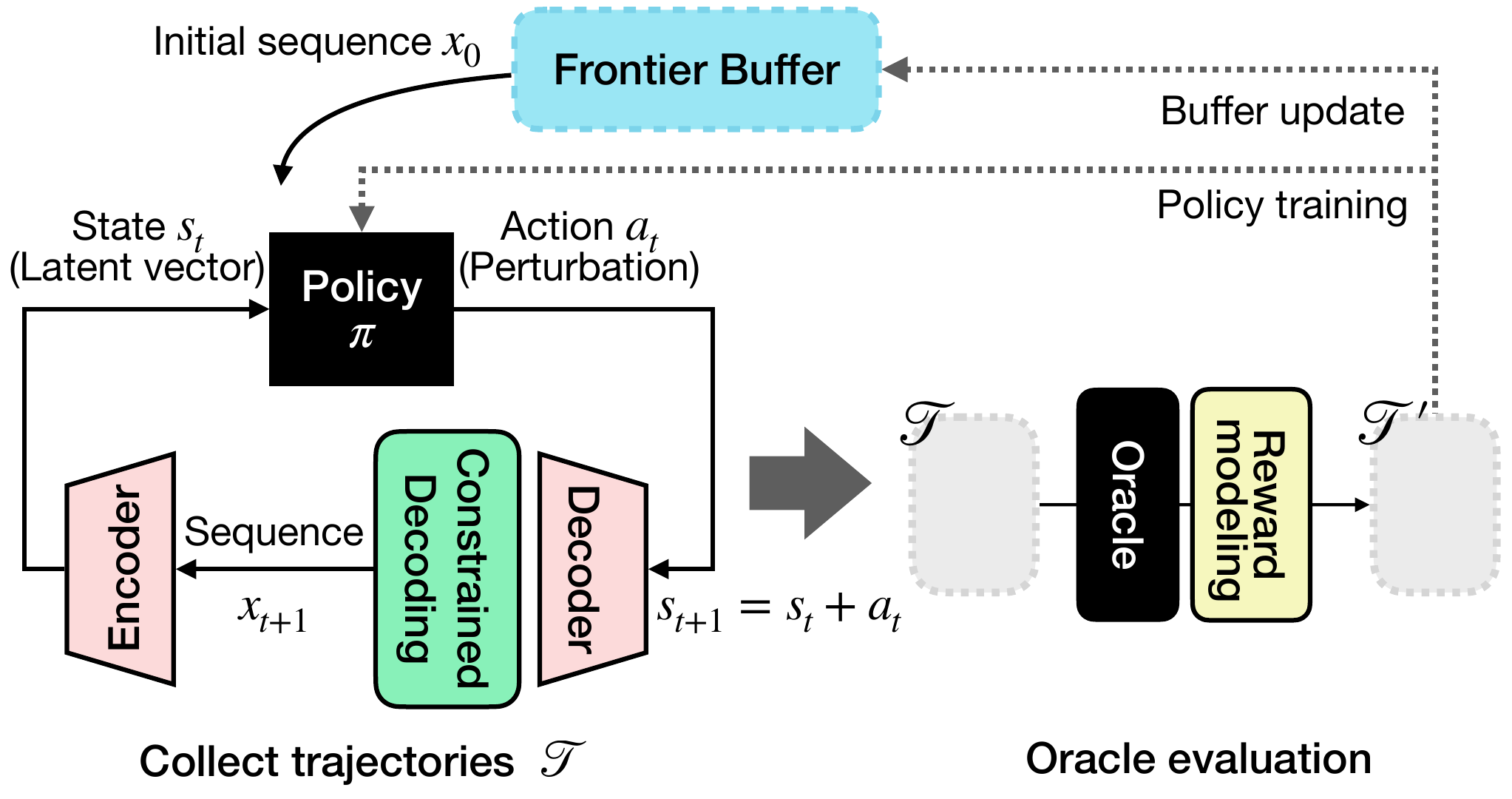}
    \vspace*{-3mm}
    \caption{\textbf{Overview of LatProtRL}. At each round, an RL policy $\pi$ acts to collect trajectories $\mathcal{T}$ for a fixed number of episodes. After $\mathcal{T}$ are collected, the reward is calculated based on the feedback from an oracle. The trajectories with calculated rewards, $\mathcal{T}'$, are used to train the policy using an on-policy RL algorithm.}
    \label{fig:overview}
\end{figure}

\subsection{Variant Encoder-Decoder (VED)}

\begin{figure*}[t!]
    \centering
    \includegraphics[width=.6\linewidth]{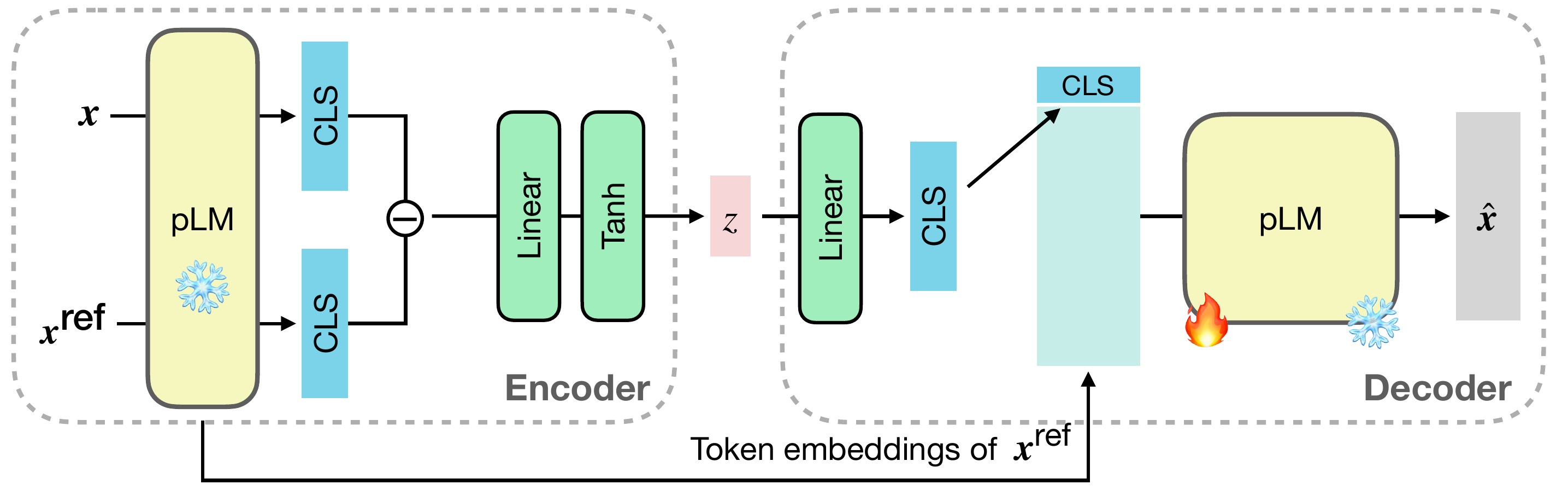}
    \vspace*{-3mm}
    \caption{\textbf{Variant Encoder-Decoder Architecture}. Given an input sequence, the encoder calculates a representation that is used by the decoder to reconstruct the original sequence. The term CLS represents the embeddings for the classification token in ESM-2.}
    \label{fig:mutation}
\end{figure*}

\looseness=-1 For successful optimization in a latent space, an informative latent space and a high-accuracy decoder are required. Additionally, the dimension of representation $R$ should ensure the effective training of the policy or generator. To address these conflicting objectives, we propose to train an VED to learn the \textit{mutation} information of a \textit{target} sequence when compared to a \textit{reference} sequence. We define the \textit{reference} $\boldsymbol{x}^\textit{ref}$ as the sequence with the minimum average distance from all the other sequences in the training dataset $\mathcal{D}$, with the distance defined as the number of mutations between two sequences. The VED architecture is shown in Figure~\ref{fig:mutation}.

\paragraph{Encoder Architecture} The encoder $\mathcal{E}_\theta$ consists of a pre-trained pLM and a dimensionality reduction layer. Given a sequence $\boldsymbol{x}$ as an input, the pLM output is given as $v \in \mathbb{R}^{(L+2) \times E_1}$, where $E_1$ is the embedding size. In ESM-2, a CLS (classification) token is prepended, and an EOS (end-of-sequence) token is appended to the sequence when generating the embeddings. We use only the CLS embedding which contains per-protein information \cite{wu2023cls}, reducing the dimensionality from $\mathbb{R}^{(L+2) \times E_1}$ to $v_\text{CLS} \in \mathbb{R}^{E_1}$. The CLS embedding is chosen over average pooling as it led to higher performance during our experiments. We subtract the CLS embedding of the mutant $v_\text{CLS}$ from the CLS embedding of the reference sequence $v_\text{CLS}^\textit{ref}$, to extract information about mutations. Given that $E_1 = 1280$, we use a single fully-connected layer to further reduce the dimension to $\mathbb{R}^R$ and apply a tanh activation function to obtain the final representation $z$.

\paragraph{Decoder Architecture} We propose the decoding architecture using a prompt tuning methodology, where we prompt the token embeddings of the reference sequence $\boldsymbol{x}^\textit{ref}$ with the mutation embedding reconstructed from the reduced representation $z$ using a fully-connected layer. Specifically, prompt embedding replaces the original CLS embedding of $\boldsymbol{x}^\textit{ref}$. The combined $(L+2, E_2)$ embedding is passed through ESM-2 attention layers. During training, the initial 4 attention layers are fine-tuned so that it can adapt to the new CLS embedding. Compared to conventional approaches that fine-tune the last layers of language models, we observed that in our case, as we change the input of the pLM, fine-tuning the initial layers leads to better performance. Finally, we reconstruct the sequence $\hat{\boldsymbol{x}}$ using a prediction head, which is not updated during fine-tuning.

\paragraph{Constrained Decoding} After each policy action, the entire amino acid sequence $\hat{\boldsymbol{x}}_{t+1}$ is reconstructed from $s_{t+1}$. Even though the proposed VED shows high accuracy (see Section~\ref{sec:results}), the decoder accuracy is still a bottleneck for small $R$ and large $L$. 
We tackle this with a constrained decoding strategy, in which we apply the top $m_\text{decode}$ mutations with respect to their predicted probabilities by the decoder. These mutations are then applied to a $\boldsymbol{x}_\textit{template}$ sequence, which is the sequence of the previous state in our optimization algorithm. 
%

\looseness-1 The VED training objective is the cross-entropy loss between $\boldsymbol{x}$ and $\hat{\boldsymbol{x}}$. We augment the training dataset $\mathcal{D}$ with random mutations, such that mutated sequence has the expected number of mutations from the original sequence equal to 3.

\subsection{Protein Fitness Optimization via Model-Based RL} 


Protein fitness landscapes can exhibit multiple peaks in which high-fitness proteins are located \citep{kauffman1989nk}, and single mutations can drastically change the fitness. Regularization methods that create a convex or smoothed landscape \citep{castro2022relso,kirjner2023optimizing} may not represent the true fitness landscape.
We formulate the problem as a Markov decision process (MDP) where the agent optimizes the sequence in a latent space through multiple timesteps of an episode traversing the fitness landscape. 
See Figure~\ref{fig:overview} and Algorithm~\ref{alg:main} for details on LatProtRL.

\begin{algorithm}[t!]
    \caption{LatProtRL algorithm}
   \label{alg:main}
    \begin{algorithmic}[1]
    \State Initialize buffer $\mathcal{B} \gets$ \Call{initialize}{$\mathcal{D}$}
    \For{$E$ rounds}
        \State Set of trajectories $\mathcal{T}\gets\{\}, N_\text{ep}\gets 0$
        \While{$N_\text{ep} < N_\text{oracle\_calls}$}
            \State Sample initial sequence $\boldsymbol{x}_0 \gets $ \Call{top}{}()
            \State Encode initial state $s_0 \gets \mathcal{E}_\theta (\boldsymbol{x}_0)$
            \State $t \gets 0,$ \texttt{done}$\gets$ \texttt{false}
            \While{not \texttt{done}}
                \State Choose action $a_t$ at state $s_t$ following $\pi$
                \State Next state $s_{t+1} \gets s_t + a_t $
                \color{Orchid}
                \State \textls[-40]{Reconstruct sequence $\boldsymbol{x}_{t+1} \gets \mathcal{R}_\phi (s_{t+1},\boldsymbol{x}_{t})$}
                \State \Comment{\textbf{Constrained decoding}}
                \color{MidnightBlue}
                \If{$d(\boldsymbol{x}_{t+1}, \boldsymbol{x}_{t})>m_\text{step}$} \texttt{v} $\gets$ \texttt{false}
                \Else \texttt{ v} $\gets$ \texttt{true}, $N_\text{ep}\gets N_\text{ep} + 1$
                \State \Comment{\textbf{Calibrating steps}}
                \EndIf 
                \color{black}
                \If{$d(\boldsymbol{x}_{t+1}, \boldsymbol{x}_0)>m_\text{total}$ or $t+1=T_{ep}$}
                \State \texttt{done}$\gets$\texttt{true} 
                \Else \texttt{ done}$\gets$\texttt{false}
                \EndIf
                \State $\mathcal{T}\gets \mathcal{T} \cup \{s_t, a_t, s_{t+1}, \boldsymbol{x}_{t+1}, $\texttt{v}, \texttt{done}$\}$
                \State $t \gets t+1$
            \EndWhile
        \EndWhile 
        \State Set of trajectories $\mathcal{T}'\gets\{\}$
        \For{$\{s_t, a_t, s_{t+1},\texttt{v}, \boldsymbol{x}_{t+1}, $\texttt{done}$\} \in \mathcal{T}$}
            \color{MidnightBlue}
            \If{\textbf{not} \texttt{v}} $r_t \gets -1$
            \color{black}
            \ElsIf{\texttt{done}}
                \State $r_t \gets q(\boldsymbol{x}_{t+1})$
                \State \textsc{update}($\boldsymbol{x}_{t+1}, r_t$)
            \Else $\text{ }r_t \gets 0$
            \EndIf
            \State $\mathcal{T}'\gets \mathcal{T}' \cup \{s_t, a_t, r_t, s_{t+1}, $\texttt{done}$\}$
        \EndFor 
        \State Update policy $\pi$ using $\mathcal{T}'$
    \EndFor
    \end{algorithmic}
\end{algorithm}

At the beginning of each episode, we sample an initial sequence $x_0 \in \mathcal{B}$ from a \textit{frontier buffer} (see Section~\ref{sec:buffer}). We map the sequence $x_0$ to a state $s_0 \gets \mathcal{E}_\theta (\boldsymbol{x}_0)$ in the representation space (state space) $\mathcal{Q}$. At each timestep $t$, the agent observes a state $s_t \in \mathcal{Q}$ and selects an action $a_t \in \mathcal{A}$ according to a policy $\pi:\mathcal{Q} \to \mathcal{A}$, where $\mathcal{A}$ is the action space. The action $a_t$ is defined as a perturbation in which a continuous value $a_{t_{j}} \in [-\delta, \delta]$ is chosen for the $j$-th dimension, and $\delta$ is a hyperparameter that is chosen based on the representation distribution. Both $s_t$ and $a_t$ have $R$ dimensions. The next state is given as $s_{t+1} = s_t + a_t$. The agent interacts with the environment until: (i) the last timestep $T_{ep}$ of an episode is reached; or
(ii) $d(\boldsymbol{x}_{t+1}, \boldsymbol{x}_0) > m_\text{total}$ where $m_\text{total}$ is the maximum number of mutations allowed per episode. The distance function $d$ is defined as the number of different amino acids between two sequences.

\begin{algorithm}[t!]
\caption{Frontier buffer operations.}
   \label{alg:buffer}
    \begin{algorithmic}
    \State Buffer $\mathcal{B} \gets \{\}$, Buffer size $S_\mathcal{B} \gets 128 $
    \State Exploration probability $\epsilon \gets 1.0$
    \State \textls[-20]{Hyperparameters for $\epsilon$ updates $T\gets 0, T_u \gets 50, \gamma \gets 0.96$}
    \Function{initialize}{$\mathcal{D}$}
    \For{$(\boldsymbol{x}, f(x)) \in$ random $S_\mathcal{B}$ elements of $\mathcal{D}$}
        \State $\mathcal{B} \gets\mathcal{B} \cup (\boldsymbol{x}, f(x), 1)$
    \EndFor 
    \EndFunction
    
    \Function{top}{}()
    \State $T\gets T+1$
    \If{$T$ is multiple of $T_u$} $\epsilon \gets \max(0.05, \gamma \cdot \epsilon)$
    \EndIf
    \State $r\gets$ random value $\in[0,1)$
    \If{$r<\epsilon$} $b\gets $ sample index from $\mathcal{B}$ where probability $p(b)\propto 1/\sqrt{\mathcal{B}[b][2]}$
    \Else $\text{ }b\gets $ sample index from $\mathcal{B}$ where probability $p(b) \propto \text{softmax} (c \cdot \mathcal{B}[b][1])$ \Comment{$c$ is a temperature}
    \EndIf
    \State $\mathcal{B}[b][2]  \gets \mathcal{B}[b][2] + 1$
    \State \Return{$\mathcal{B}[b][0]$}
    \EndFunction
    \Function{update}{$\boldsymbol{x}, f(x)$}
    \If{$\boldsymbol{x} \notin \mathcal{B}$}
    \State $b_\text{min} \gets$ index of element of minimum fitness in $\mathcal{B}$
    \If{$f(x)>\mathcal{B}[b_\text{min}][1]$}
    \Comment{Replace element}
    \State $\mathcal{B}[b_\text{min}] \gets (\boldsymbol{x}, f(x), 1)$
    \EndIf
    \EndIf
    \EndFunction
    \end{algorithmic}
\end{algorithm}

\textbf{Calibrating Steps}. 
\looseness=-1 In our formulation, it is desired to optimize through multiple steps, since it enables the value function to learn states in high-fitness regions. Additionally, we want actions that avoid deviating very far from the initial sequences, since these might lead to low-fitness regions of the fitness landscape. As a way to calibrate the policy actions, we give negative feedback for actions leading to a high number of mutations per step. The variable \texttt{v} is set to \texttt{false} if $d(\boldsymbol{x}_{t+1}, \boldsymbol{x}_t) > m_\text{step}$ and stored in trajectory. During the reward calculation step, if \texttt{v} is \texttt{false} the oracle is not called and a negative reward is assigned to the policy. The effect of calibrating steps is studied in Section~\ref{sec:ablations}.




\textbf{Reward Function}. At each timestep of an episode, a 6-tuple $(s_t, a_t, s_{t+1}, \boldsymbol{x}_{t+1},$\texttt{v},\texttt{done}) without reward $r_t$ is stored in $\mathcal{T}$. The reward is calculated only after all the trajectories are collected using the current policy $\pi$. The reward is defined as a sparse reward where we only evaluate the optimized sequence at the last timestep using the oracle. The reward function is shown in lines 24-27 of Algorithm~\ref{alg:main}.  The trajectories are updated using the computed rewards and used to train the policy using Proximal Policy Optimization (PPO) \cite{schulman2017proximal}.  We finish each optimization round when the number of episodes $N_\text{ep}$ reaches the number of oracle calls set per round, $N_\text{oracle\_calls}$. When using in vitro oracles, $N_\text{oracle\_calls}$ is set based on the budget for wet lab experiments.



\subsection{Frontier Buffer}
\label{sec:buffer}

We propose a \textit{frontier buffer}\footnote{This concept is different from the \textit{replay buffer} concept that stores transitions in off-policy RL.} inspired by Go-Explore~\cite{ecoffet2021first}, which has shown that storing and starting from the states that led to high reward enhance the performance of RL algorithms tackling environments in which exploration is challenging. Algorithm~\ref{alg:buffer} details the buffer operations. 
The frontier buffer is a set of 3-tuple $(\boldsymbol{x}, f$,\texttt{visit}$)$ with fixed size $S_\mathcal{B}$, where $f$ is the fitness of sequence $\boldsymbol{x}$, and \texttt{visit} is the number of episodes in which $\boldsymbol{x}$ was sampled as the initial state. 
The buffer is initialized by sampling $S_\mathcal{B}$ sequences from $\mathcal{D}$, with visit count set to 1. At the beginning of an episode, the initial sequence $x_0$ is sampled from the buffer. We employ an $\epsilon$-greedy formulation to balance exploitation and exploration. 
We replace the sequence with the lowest fitness in the buffer with a new sequence if it has higher fitness at the end of each round. The buffer does not allow repeated sequences by design. We use the sequences in the buffer after the final round to measure the performance of our method.

%


\begin{table}[t!]
    \centering
    \begin{tabular}{cccc}
    \toprule
        Dataset & $\mathcal{D}^*$ & \textit{medium} & \textit{hard} \\
    \midrule
        AAV & 0.466 & 0.376 & 0.326 \\
        GFP & 0.738 & 0.232 & 0.092 \\
    \bottomrule
    \end{tabular}
    \caption{\textbf{Median fitness }of $\mathcal{D}^*$ and the top 128 sequences of the \textit{medium} and \textit{hard} tasks subsets.}
    \label{tab:stats}
\end{table}

\section{Results}
\label{sec:results}

\subsection{Experiment Setup}

\paragraph{Datasets and Oracles}

\begin{table*}[t!]
    \centering
    \setlength{\tabcolsep}{3pt}
    \begin{tabular}{lcccc|cccc}
    \toprule
         & \multicolumn{4}{c}{AAV \textit{medium} task} & \multicolumn{4}{c}{AAV \textit{hard} task} \\
         \cmidrule(lr){2-5} \cmidrule(lr){6-9}
         Method & Fitness $\uparrow$ & Diversity & $d_\text{init}$ & $d_\text{high}$ & Fitness $\uparrow$ & Diversity & $d_\text{init}$ & $d_\text{high}$ \\
    \midrule
        PEX & 0.64 (0.0) & 5.4 (0.5) & 5.2 (0.4) & 4.0 (0.0) & 0.62 (0.0) & 5.8 (0.7) & 6.2 (1.0) & 6.0 (0.0) \\
        AdaLead & \textbf{0.74} (0.0) & 3.8 (0.4) & 7.4 (1.0) & 4.4 (1.0) & \textbf{0.72} (0.0) & 3.2 (0.4) & 8.2 (1.2) & 6.4 (1.0) \\
        BO & 0.59 (0.0) & 8.8 (0.4) & 10 (0.5) & 9.6 (1.0) & 0.61 (0.0) & 8.6 (0.5) & 11 (0.5) & 8.6 (1.0) \\
        CMAES & 0.03 (0.0) & 21 (0.5) & 20 (0.9) & 19 (0.7)& 0.04 (0.0) & 21 (0.5) & 20 (0.5) & 19 (1.0)\\
        CMAES-VED & 0.61 (0.0) & 4.8 (0.4) & 9.1 (0.2) & 5.9 (0.9) & 0.50 (0.0) & 5.0 (0.0) & 9.6 (0.5) & 5.4 (1.3) \\
        \textbf{LatProtRL} & \textbf{0.71} (0.0) & 5.4 (0.9) & 6.1 (0.2) & \textbf{2.4} (0.5) & \textbf{0.66} (0.0) & 6.0 (1.2) & 7.0 (0.0) & \textbf{2.0} (0.0) \\
    \rowcolor{lightgray!20}
        -- w/o buffer & 0.49 (0.0) & 11 (0.0) & 4.0 (0.0) & 6.8 (0.5) & 0.42 (0.0) & 13 (0.6) & 4.3 (0.6) & 8.3 (0.6) \\
    \rowcolor{lightgray!20}
        -- w/o calibrating steps & 0.62 (0.0)& 7.4 (0.9) & 6.0 (0.0) & 4.5 (0.6) & 0.64 (0.0) & 7.4 (0.9) & 7.2 (0.4) & 5.4 (0.6) \\
    \rowcolor{lightgray!20}
        -- PPO (Lat/Mut)  & 0.57 (0.0) & 5.7 (1.1) & 6.7 (1.1) & 9.0 (1.0) & 0.59 (0.0) & 6.2 (1.3) & 7.0 (1.0) & 8.6 (0.9) \\
    \rowcolor{lightgray!20}
        -- PPO (Seq/Mut) & 0.55 (0.0) & 7.0 (0.8) & 7.6 (1.0) & 9.3 (0.5) & 0.60 (0.0) & 4.8 (0.8) & 6.6 (0.5) & 7.0 (0.7) \\
    \toprule
        & \multicolumn{4}{c}{GFP \textit{medium} task} & \multicolumn{4}{c}{GFP \textit{hard} task} \\
        \cmidrule(lr){2-5} \cmidrule(lr){6-9}
         Method & Fitness $\uparrow$ & Diversity & $d_\text{init}$ & $d_\text{high}$ & Fitness $\uparrow$ & Diversity & $d_\text{init}$ & $d_\text{high}$ \\
    \midrule
        PEX & 0.60 (0.1) & 7.2 (1.0) & 8.0 (2.2) & 11 (2.0) & 0.52 (0.1) & 6.8 (0.7) & 13 (2.3) & 13 (3.0) \\
        AdaLead & \textbf{0.93} (0.0) & 5.0 (0.6) & 10 (1.9) & 14 (2.0) & \textbf{0.75} (0.1) & 3.8 (1.3) & 16 (1.5) & 19 (1.0) \\
        BO & 0.25 (0.1) & 31 (12) & 41 (26) & 46 (25) & 0.33 (0.1) & 38 (6.4) & 58 (20) & 66 (20) \\
        CMAES & -0.05 (0.0) & 177 (4.2) & 168 (5.7) & 167 (5.6) & -0.03 (0.0) & 176 (0.9) & 220 (4.2) & 220 (4.2) \\
        CMAES-VED & 0.57 (0.1) & 5.0 (1.0) & 9.0 (0.0) & 3.6 (0.5) & 0.44 (0.1) & 4.8 (0.4) & 9.2 (0.8) & 3.6 (0.9) \\
        \textbf{LatProtRL} & \textbf{0.93} (0.0) & 4.6 (0.5) & 6.0 (0.0) & \textbf{1.0} (0.0) & \textbf{0.85} (0.0) & 4.8 (0.5) & 7.0 (0.0) & \textbf{2.0} (0.0) \\
         \rowcolor{lightgray!20}
            -- w/o buffer & 0.64 (0.0) & 8.8 (0.4) & 3.0 (0.0) & 3.0 (0.0) & 0.49 (0.0) & 8.8 (0.4) & 3.0 (0.0) & 3.0 (0.0) \\
        \rowcolor{lightgray!20}
            -- w/o calibrating steps & 0.92 (0.0) & 4.0 (0.0) & 6.0 (0.0) & 1.1 (0.2) & 0.82 (0.0) & 4.0 (0.0) & 6.0 (0.0) & 2.0 (0.0) \\
        \rowcolor{lightgray!20}
            -- PPO (Lat/Mut) & 0.74 (0.0) & 11 (5.7) & 10 (1.8) & 14 (1.2) & 0.64 (0.0) & 6.6 (0.9) & 11 (1.5) & 16 (1.4) \\
        \rowcolor{lightgray!20}
            -- PPO (Seq/Mut) & 0.77 (0.0) & 7.8 (2.0) & 11 (1.9) & 13 (1.6) & 0.62 (0.0) & 10 (3.2) & 12 (2.5) & 15 (6.4)\\
        \bottomrule
    \end{tabular}
    \caption{\textbf{AAV and GFP optimization results} for LatProtRL and baseline methods. Shaded rows indicate the result of ablation studies in Section~\ref{sec:ablations}. The standard deviation of 5 runs with different random seeds is indicated in parentheses.}
    \label{tab:main}
\end{table*}

Following \citet{kirjner2023optimizing}, we evaluate our method in two proteins, GFP and AAV.
The length $L$ is 237 for GFP and 28 for the functional segment of AAV. The $\mathcal{D}^*$ for GFP \citep{sarkisyan2016local} contains 54,025 mutant sequences, with log-fluorescence intensity associated with each sequence. The $\mathcal{D}^*$ for AAV \citep{bryant2021deep} contains 44,156 sequences associated with its ability to package a DNA payload. The fitness values in both datasets are min-max normalized. We use the \textit{medium} and \textit{hard} level benchmarks proposed by \citet{kirjner2023optimizing} to sample $\mathcal{D}$ for optimization. The \textit{medium} task sample sequences with fitness ranging from the 20-40th percentile while the \textit{hard} task sample sequences with fitness ranging from the 10-30th percentile of $\mathcal{D}^*$. All methods including the baselines start optimization from the top 128 sequences of $\mathcal{D}$ of each task. See Table~\ref{tab:stats} for the statistics of $\mathcal{D}^*$ and $\mathcal{D}$. 
For our experiments, we use the checkpoints of the oracles and predictors in \citet{kirjner2023optimizing}.
The number of rounds $E$ and the number of oracle calls $N_\text{oracle\_calls}$ are limited in the active learning setting. We set $E=15$ and $N_\text{oracle\_calls}=256$.
For the experiments in Section~\ref{sec:predictor}, a predictor is used to calculate the reward function, and the oracle is used only for evaluation. In this case, we assume that the computational costs and inference time allow the running of multiple optimization steps and predictor calls.


\paragraph{Baselines} For the active learning setting, we compare with representative multi-round optimization methods that assume black-box oracle evaluations: Bayesian optimization, implemented as in the FLEXS benchmark \cite{sinai2020adalead}, and three evolutionary algorithms: AdaLead \cite{sinai2020adalead}, PEX \cite{anand2022protein}, and covariance matrix adaptation evolution strategy (CMAES) using one-hot encoding. We also compare with CMAES using our VED for encoding, termed CMAES-VED. In Section~\ref{sec:predictor}, we additionally compare with representative single-round optimization methods that assume query to fitness predictors: GFlowNets (GFN\_AL) \cite{jain2022biological} and Gibbs sampling (GGS) \cite{kirjner2023optimizing}. 



\paragraph{Evaluation Metrics}

We use four evaluation metrics: fitness, diversity, distance from the set of initial sequences ($d_\text{init}$), and high fitness sequences ($d_\text{high}$). While high fitness and diversity are desired, distance metrics are not deterministic, and rather provide a higher-level view of the position of optimized sequences on the fitness landscape.
Let the optimized sequences $\mathcal{G}^* = \{g^*_1, \cdots, g^*_K\}$. Fitness is defined as the median of the evaluated fitness of $K=128$ generated sequences.
Diversity is defined as the median of the distances between every pair of sequences in $\mathcal{G}^*$. The variables $d_\text{init}$ and $d_\text{high}$ are defined as the median of the minimum distance from each sequence in $\mathcal{G}^*$ to $\mathcal{D}$ and to the top 10\% fitness sequences of $\mathcal{D}^*$, respectively.

\paragraph{Implementation details}
The latent representation space $R=32$ for GFP and $R=16$ for AAV. For the policy, the perturbation magnitude $\delta$ of action vector is set to $0.3$ for GFP and to $0.1$ for AAV. The episode length $T_\text{ep}$ is set to 6 for GFP and to 4 for AAV. The value for $m_\text{step}$ is set to 3 and for $m_\text{total}$ is set to 15. The exploration buffer size $S_B$ is set to 128. The constrained decoding term $m_\text{decode}$ is set to 18 for GFP and 8 for AAV, considering their length $L$. 


\subsection{Fitness Optimization}

\begin{figure*}[ht!]
    \centering
    \includegraphics[width=0.89\linewidth]{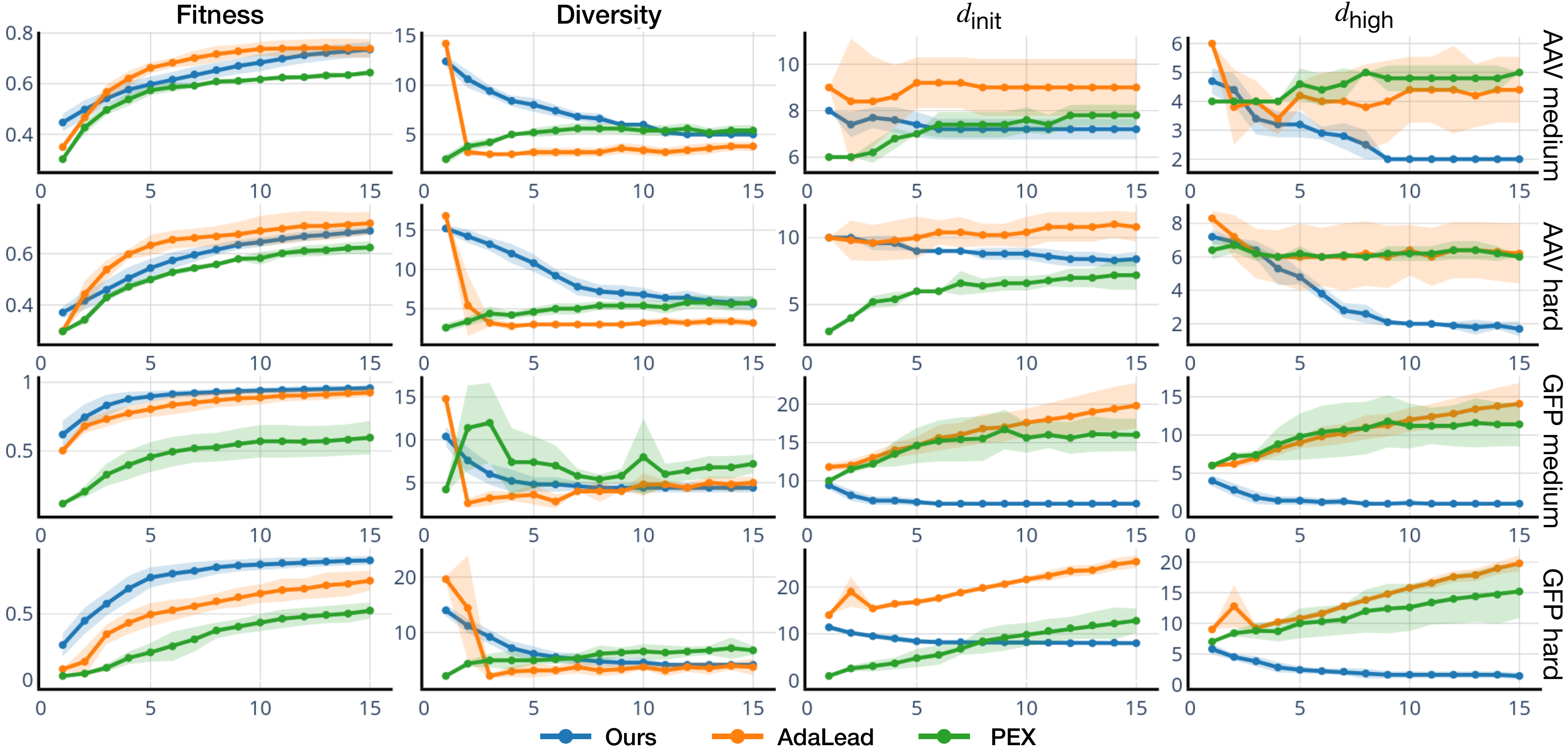}
    \caption{\textbf{Evaluation metric by optimization round} for LatProtRL, AdaLead and PEX. Shaded regions indicate the standard deviation of 5 runs. The x-axis indicates the number of rounds.}
    \label{fig:policy-training}
\end{figure*}

We report the mean and standard deviation of the evaluation metrics of 5 runs with different random seeds in Table~\ref{tab:main}. LatProtRL outperforms or shows comparable results with baseline methods. The 90th percentile normalized fitness is 0.64 for AAV and 0.86 for GFP in the experimental datasets, suggesting that our method successfully optimizes all tasks. For AAV optimization, AdaLead shows the highest fitness, whereas LatProtRL shows the higher diversity when compared to the evolutionary algorithm baselines. LatProtRL achieves the highest fitness while achieving comparable diversity for GFP \textit{medium} and \textit{hard} tasks. In contrast, CMAES fails to optimize both sequences with one-hot encoding, but achieves comparable results to PEX using our VED representation.

\begin{figure*}[ht!]
    \centering
    \includegraphics[width=\linewidth]{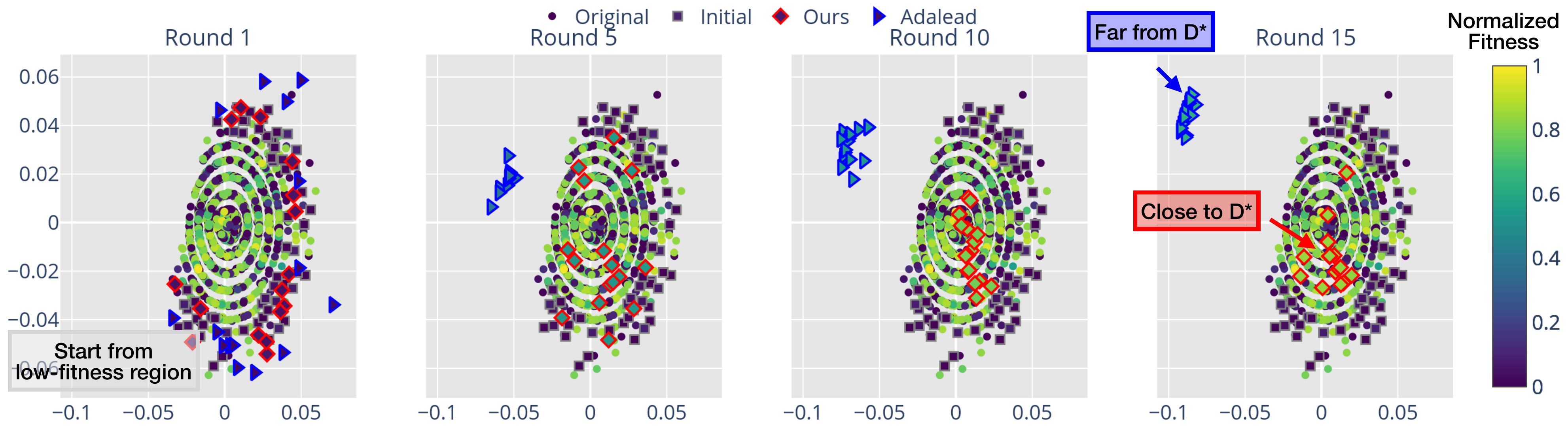}
    \caption{\textbf{Optimization trajectories of LatProtRL and AdaLead} in GFP \textit{hard} task. The ``original" term indicates the experimental rugged fitness landscape, exhibiting several local peaks. The x- and y-axis are obtained by multidimensional scaling (MDS) \citep{kruskal1964nonmetric} of pairwise distances of 2500 sequences sampled from $\mathcal{D}^*$ with 16 optimized sequences from LatProtRL and AdaLead at each round. We chose the median of 16 sequences but observed a similar tendency for the top 16 sequences. AdaLead generates improved sequences at each round but farther from the experimental data distribution and with fitness values lower when compared to high-fitness sequences (See \textcolor{blue}{$\triangleright$} markers at Round 15). LatProtRL generates sequences closer to high-fitness sequences in the data distribution (See \textcolor{red}{$\Diamond$} markers at Round 15) while also escaping local optima. 
    }
    \label{fig:optimization-trajectories}
\end{figure*}

LatProtRL is more successful in GFP when compared to AAV, and when initial sequences $\mathcal{D}$ exhibit lower fitness. We demonstrate that the reason for this is related to the difference in the characteristics of both dataset landscapes. The GFP dataset~\cite{sarkisyan2016local} is collected by random mutagenesis, in which initial sequences exhibit very low ($<0.1$) fitness and are far from the region with high-fitness sequences (See Figure~\ref{fig:optimization-trajectories}). On the other hand, the AAV dataset~\cite{bryant2021deep} uses an additive fitness predictor to sample sequences with predicted high fitness for experiments. For AAV, 62.5\% of random mutants with more than 6 mutations have high fitness. The results for GFP indicates that LatProtRL can be particularly useful for rugged fitness landscapes given its MDP formulation.

\paragraph{Evaluation by Optimization Round}
Figure~\ref{fig:policy-training} shows the evaluation results after each round of optimization. The policy effectively improves the performance of the generated sequences over rounds. Also, the proposed method alleviates the decrease in diversity of the optimized sequence over rounds when compared to AdaLead. 

\paragraph{Analysis of Distance Metrics}
As shown in Figure~\ref{fig:policy-training}, LatProtRL decreases the distance between optimized sequences and the top 10\% sequences of $\mathcal{D}^*$, reaching a median distance of 1.0 and 2.0 for GFP \textit{medium} and \textit{hard}, respectively. Remarkably, these highly-fit sequences were never used in the VED's training or optimization process.
This strength is unique to LatProtRL and is not observed in other baselines.
Given that the GFP landscape is highly centralized around the wild-type, and that only 0.4\% of GFP mutants with over 10 mutations from the wild-type show fitness above $0.5$, with the highest observed fitness being $0.79$, a high $d_\text{high}$ might suggest false positives. Thus, LatProtRL reaching low $d_\text{high}$ indicates the ability to generate sequences at the high-fitness regions of the experimental data even starting from low-fitness regions far from the wild-type sequence.

\subsection{Optimization Using a Fitness Predictor}
\label{sec:predictor}

\begin{table}[t!]
    \centering
    \setlength{\tabcolsep}{3pt}
    \begin{tabular}{lcccccc}
    \toprule
        & \multicolumn{3}{c}{AAV \textit{medium}} &\multicolumn{3}{c}{AAV \textit{hard}} \\
        \cmidrule(lr){2-4} \cmidrule(lr){5-7}
         Method & Fitness & Div. & $\text{d}_\text{init}$ & Fitness & Div. & $\text{d}_\text{init}$\\
    \midrule
        GFN-AL & 0.18 (0.1) & 9.6 & 19 & 0.10 (0.1) & 9.5 & 19 \\ 
        CbAS & 0.47 (0.1) & 8.8 & 5.3 & 0.40 (0.0) & 12 & 7.0 \\
        AdaLead & 0.43 (0.0) & 3.8 & 2.0 & 0.44 (0.0) & 2.9 & 2.0 \\ 
        GGS & \textbf{0.51} (0.0) & 4.0 & 5.4 & \textbf{0.60} (0.0) & 4.5 & 7.0 \\
        \textbf{LatProtRL} & \textbf{0.57} (0.0) & 3.0 & 5.0 & \textbf{0.57} (0.0) & 3.0 & 7.0 \\
    \toprule
        & \multicolumn{3}{c}{GFP \textit{medium}} & \multicolumn{3}{c}{GFP \textit{hard}}\\
        \cmidrule(lr){2-4} \cmidrule(lr){5-7}
         Method & Fitness & Div. & $\text{d}_\text{init}$ & Fitness & Div. & $\text{d}_\text{init}$ \\
    \midrule
        GFN-AL & 0.15 (0.1) & 16 & 213 & 0.16 (0.2) & 22 & 215 \\ 
        CbAS & 0.66 (0.1) & 3.8 & 5.0 & 0.57 (0.0) & 4.2 & 6.3 \\
        AdaLead & 0.59 (0.0) & 5.5 & 2.0 & 0.39 (0.0) & 3.5 & 2.0 \\ 
        GGS & \textbf{0.76} (0.0) & 3.7 & 5.0 & \textbf{0.74} (0.0) & 3.6 & 8.0 \\
        \textbf{LatProtRL} & \textbf{0.81} (0.0) & 3.0 & 5.0 & \textbf{0.75} (0.0) & 3.0 & 7.0 \\
    \bottomrule
    \end{tabular}
    \caption{\textbf{Single-round optimization results} using predictor trained on $\mathcal{D}$. We report standard deviation for the fitness values over 5 runs with different seeds in parentheses.}
    \label{tab:opt_predictor}
\end{table}

We also investigate a scenario in which the optimization is guided by a ``known'' fitness predictor, assuming it can be trained accurately using $\mathcal{D}$.
This predictor approximates the oracle and substitutes it for the reward calculation as presented in Figure~\ref{fig:overview}. The oracle is used just for the final evaluation. The optimization results are shown in Table~\ref{tab:opt_predictor}. 
LatProtRL achieves the highest performance for the AAV \textit{medium} and GFP, while being the second best for the AAV \textit{hard} task. 
When compared to AdaLead, GGS and LatProtRL incorporate exploration methodologies in the optimization framework that leads to generating sequences farther from the initial sequences, \textit{i.e.} higher $d_{\text{init}}$. 
Notably, compared to GGS, our method does not require a differentiable predictor for sampling optimized sequences. Compared to the performance of methods using the oracle in Table~\ref{tab:main}, using the predictor as a surrogate model leads to lower fitness values for the generated sequences. 


In Appendix~\ref{app:double}, we provide another practical setup for LatProtRL. We use the model in a double-loop optimization setting, with an in silico predictor serving as a reward function between rounds of black-box evaluation. Such a setting can be used when the number of rounds is limited.

\begin{table*}[ht!]
    \centering
    \begin{tabular}{lccccc}
    \toprule
        Candidate & in silico oracle score &  $d_\text{high}$ & $d_\text{init}$ & pLDDT & Fluorescence intensity ($\times 10^7$) \\
    \midrule
        Wild-type & 0.9396 & 0 & 6 & 95.80 & 3.15 ± 0.07 \\
        adalead\_1 & 1.0360 & 7 & 12 & 95.80 & 4.16 ± 0.07 \\
        adalead\_12 & 1.0028 & 11 & 16 & 95.23 & 2.72 ± 0.44 \\
        \textbf{latprotrl\_12} & 1.0170 & 1 & 7 & \textbf{96.02} & \textbf{5.00} ± 0.16\\
        \textbf{latprotrl\_16} & 1.0156 & 3 & 8 & \textbf{96.03} & \textbf{5.01} ± 0.12 \\
    \bottomrule
    \end{tabular}
    \caption{\textbf{Statistics of the candidates}. The variable $d_\text{high}$ denotes the \underline{minimum} distance to the top 10\% functional sequences of $\mathcal{D}^*$. The variable $d_\text{init}$ denotes the \underline{median} distance to the initial dataset $\mathcal{D}$ provided to the policy. The fluorescence intensity shows mean and standard deviation from $n=3$ independent experiments.}
    \label{tab:candidates}
\end{table*}

\subsection{Essential Components in RL Modeling}
\label{sec:ablations}

\begin{figure}[t!]
    \centering
    \includegraphics[width=0.95\columnwidth]{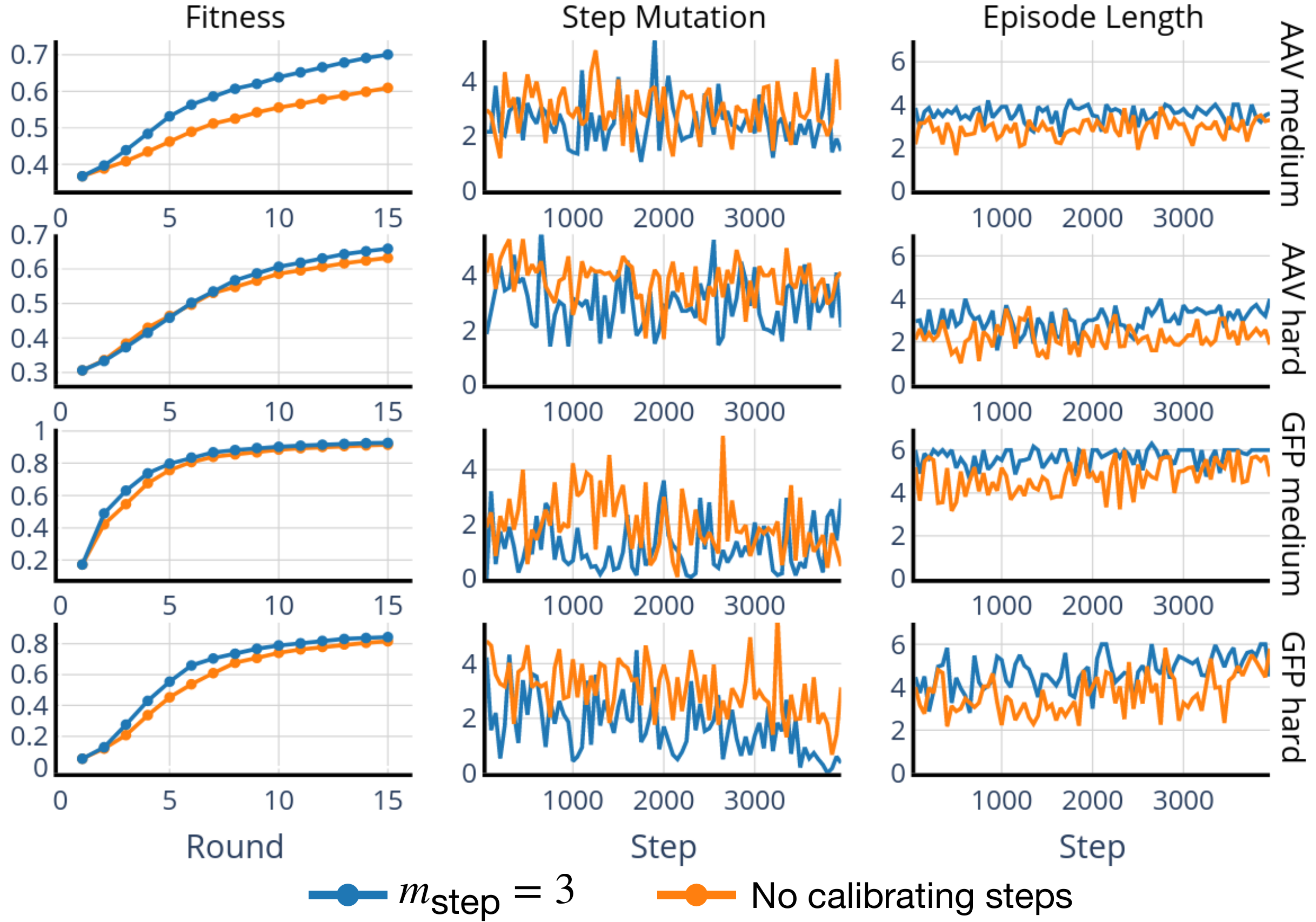}
    \caption{\textbf{Effect of the calibrating steps} to fitness and episode length. Calibrating steps allow the policy to learn actions leading to less than $m_{\text{step}}$ mutations and increasing the length of the episodes during training.}
    \label{fig:calibration-steps}
\end{figure}

\looseness=-1 We now examine the effect of different model components. First, Table~\ref{tab:main} shows the effects of training PPO with different state and action modeling. We compare two possible state modeling options: the representation obtained by the VED (Lat) and the one-hot encoded input sequence (Seq). For the action modeling we compare a perturbation in the representation (Lat) and mutation at a position in the sequence (Mut). The latent space modeling of LatProtRL leads to higher performance compared to the other modeling options. 
Second, we show the effects of the frontier buffer in 
Table~\ref{tab:main}. Without the buffer, the performance drops significantly and the RL policy needs longer episodes and leads to unstable training under a multi-round optimization setting.
%
Third, we show the effects of the calibrating steps in 
Figure~\ref{fig:calibration-steps}.
The policy learns to take actions within the maximum number of steps $m_\text{steps}$ allowed per timestep. Additionally, using the calibration steps leads to longer episode lengths over time for GFP, ultimately leading to a performance increase.

\subsection{Impact of MDP Formulation}

We discuss the impact of the MDP formulation in sampling and optimization by comparing to other top-performing methods. In \citet{sinai2020adalead,kirjner2023optimizing} the mutation rate is constrained to $1/L$ per timestep, while in our current formulation, the number of mutations is constrained by the $m_\text{decode}$, set to 12 and 8, for GFP and AAV, respectively. Additionally, \citet{sinai2020adalead} apply greedy search and \citet{kirjner2023optimizing} apply Gibbs sampling at each timestep, while we use a frontier buffer and an RL formulation maximizing a sparse reward that is only given by evaluating the sequence proposed at the last timestep. In this way, we proactively avoid local optima in the optimization framework. We illustrate this ability in Figure~\ref{fig:optimization-trajectories} and Appendix~\ref{app:localoptima} with the trajectories obtained by LatProtRL and AdaLead.
LatProtRL can avoid local optima and generate designs closer to the experimental data. In contrast, AdaLead's designs are located far from the experimental data.


\subsection{In Vitro Validation}

\begin{figure}[t!]
    \centering
    \includegraphics[width=.93\linewidth]{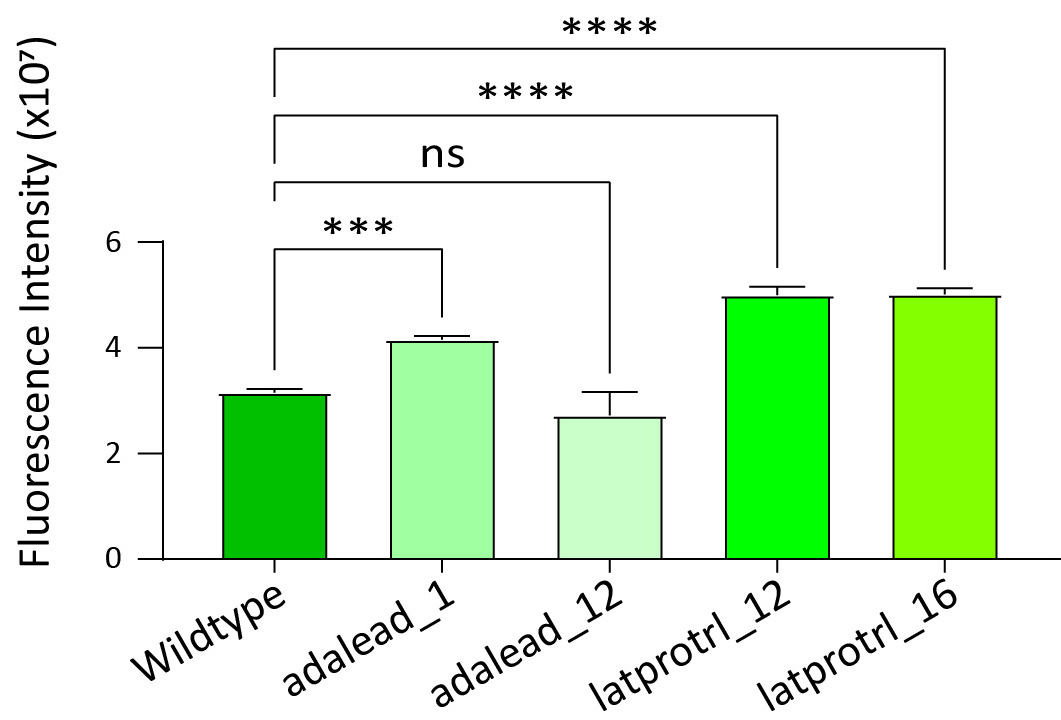}
    \caption{\textbf{The GFP fluorescence intensity} (Excitation: 485nm and emission: 535nm) \textbf{of each variant} (20$\mu$M, 0.1ml). The data from independent experiments ($n=3$) were analyzed and expressed as mean±SD (0.01$<$*P$<$0.1, 0.001$<$**P$<$0.01, 0.0001$<$***P$<$0.001, ****P$<$0.0001 vs. control). P values by one-way ANOVA test followed by Dunnett’s multiple comparisons test. \textbf{ns}, not significant.}
    \label{fig:invitro}
\end{figure}

We conducted an in vitro assessment to further support the strength of LatProtRL proposing variants close to the experimental distribution. We analyzed the 256 candidates generated after the end of round $E$=15 by LatProtRL and by AdaLead for the GFP \textit{medium} task. We selected top 30 sequences ranked by the in silico oracle. We predicted the structures using AlphaFold2 (AF2)~\cite{jumper2021highly} for the top 30 sequences of LatProtRL, the top 30 sequences of AdaLead, and the wild-type sequence. We select the top 2 sequences for each model by pLDDT values. All four selected sequences and wild-type were expressed and purified successfully. The details on the purification and fluorescence measurement of the variants are presented in Appendix~\ref{app:invitro}. The statistics of candidates including calculated fluorescence intensities are shown in Table~\ref{tab:candidates} and Figure~\ref{fig:invitro}. 

LatProtRL designs achieve the highest fluorescence intensity, with both designs achieving intensity higher than the wild-type sequence. For AdaLead, only one design achieved an intensity higher than the wild-type, even though both candidates are predicted to have higher fitness in silico. Analysis of the mutations proposed by each method when compared to the wild-type sequence are detailed in Appendix~\ref{app:seqalign}.
 


\section{Conclusion}
This paper addressed protein fitness optimization in an active learning setting, starting with low-fitness sequences. We modeled the problem as an MDP to maximize future rewards, which allows an efficient exploration of the landscape and escape from the local optima. Our framework, named LatProtRL, uses an RL policy to traverse a latent space learned by a proposed variant encoder-decoder. LatProtRL is competitive or outperformed other baseline methods for two fitness optimization benchmarks, GFP and AAV. Our results show that sequences generated by LatProtRL reach high-fitness regions of the experimental data, demonstrating its potential to be extended to lab-in-the-loop scenarios. We anticipate our research to impact real-world protein design tasks involving in vitro experiments, such as enhancing antibody affinity or protein stability. Future research directions include combining LatProtRL with the feedback provided by AlphaFold2 and extending the proposed VED to accommodate the insertion and deletion of amino acids.


\clearpage
\section*{Acknowledgements}

This work was supported by the Institute for Basic Science (IBS-R029-C2, IBS-R030-C1), Republic of Korea. We sincerely thank Jeongwon Yun for running the in vitro experiments used in the validation of the proposed method which greatly improved the quality of this work. 



\section*{Impact Statement}

This research offers novel protein engineering methods to the AI community and can help improve drug discovery and pandemic readiness. We also stress that this technology can be abused and pose biosecurity threats in making harmful substances. This worry is not unique to our research but common to all research that use AI in protein engineering.

\bibliography{reference}
\bibliographystyle{icml2024}

\newpage
\appendix
\onecolumn

\label{sec:appendix}

\section{In Vitro Evaluation}


\subsection{Protein purification and fluorescence measurement of GFP variants}
\label{app:invitro}
The full-length of GFP variants followed by a 6X-His tag and a stop codon were cloned into the NdeI and XhoI sites of the pET29b vector (69872, Novagen) and transformed in E. coli BL21 (DE3) (CP111, Enzynomics). Cells were grown at 37°C in LB broth with 0.05mg/mL kanamycin to an OD600 of 0.6. Protein expression was induced by 0.4mM IPTG (isopropyl beta-d-1-thiogalactopyranoside) and incubated for 3h at 37°C. The proteins were purified from cell lysate through affinity chromatography using Ni-NTA agarose affinity column (30210, QIAGEN). The finally purified protein exists in a solution state in DPBS (pH7.5) buffer. The green fluorescence intensity (excitation: 485nm and emission: 535nm) of each variant (20$\mu$M, 0.1mL) were measured using SpectraMax® iD5 (using SoftMax Pro 7.1.2 software).

\subsection{Sequence alignment}

\label{app:seqalign}
\begin{figure}[h!]
    \centering
    \includegraphics[width=0.8\linewidth]{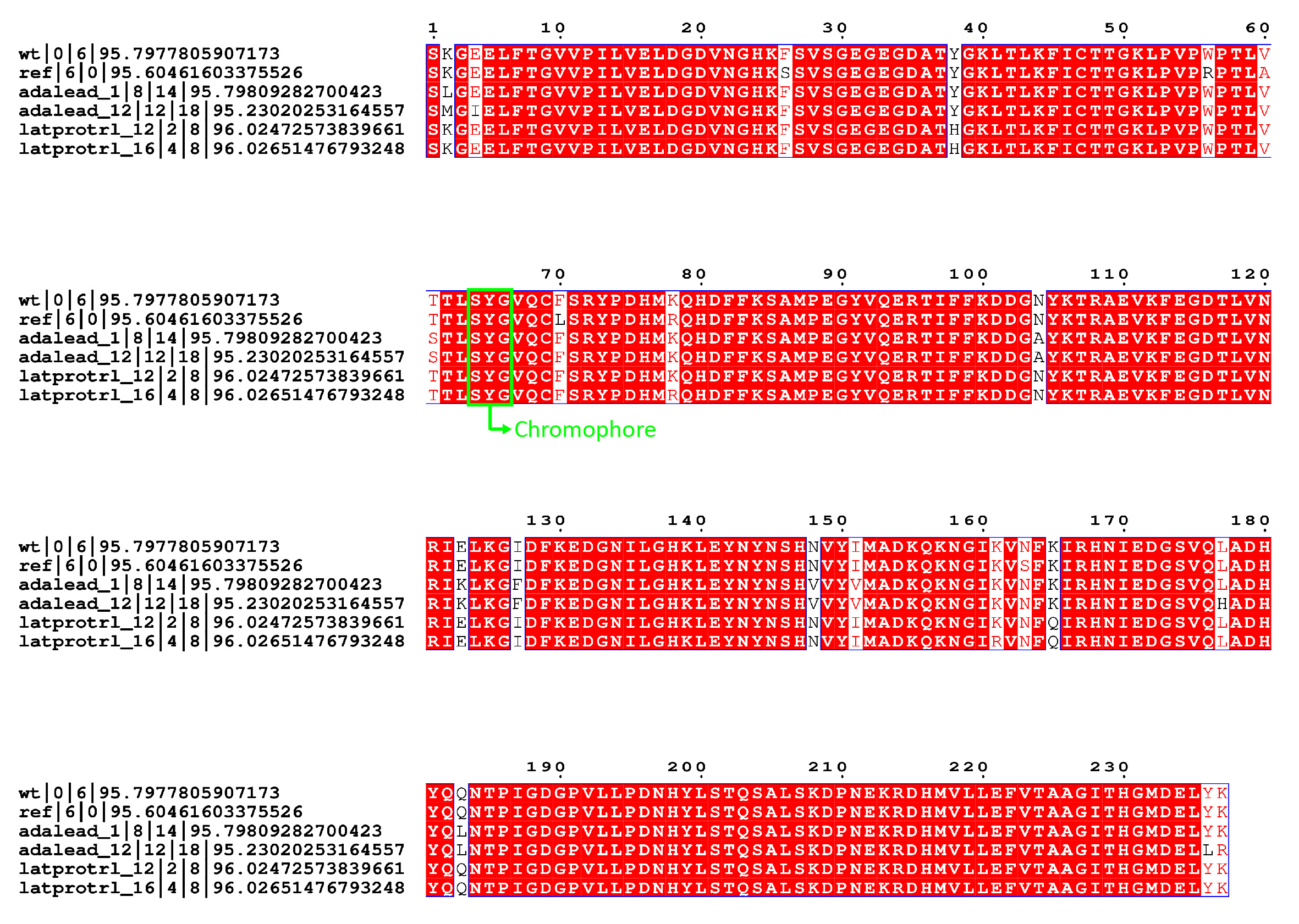}
    \caption{\textbf{Sequence alignment} for the GFP wild-type, the low-functional reference, and the top 2 designs by LatProtRL and AdaLead. For all designs, the residues in the chromophore region are kept. The label of sequence includes name, number of mutations from wild-type, number for mutations from \textit{medium} task reference sequence, and AF2 pLDDT score delimited by vertical line.}
    \label{fig:seqalign}
\end{figure}

\clearpage 

\section{Hyperparameter Search and Guidelines}

In this section, we discuss the hyperparameters of LatProtRL and provide some guidelines for the tuning and choice for different protein families. Here, we discuss the following hyperparameters: $R$, $\delta$, $m_\text{step}$, $T_\text{ep}$, $m_\text{total}$, and $m_\text{decode}$. First, we explain two important parameters in depth: $R$ (dimension of latent space) and $\delta$ (maximum magnitude of the perturbation).

We chose the value of $R$ that leads to higher decoder accuracy. In our hyperparameter search we set $R=8,16,32,64$. Policy training was conducted with $R=16$ or $R=32$ and showed no significant differences between the two. Therefore, we speculate that VED performance could be the sole factor in hyperparameter searches in resource-constrained settings. 

We use a tanh activation function as the last layer of the encoder, which inherently limits the range of $\delta$. 
We show ablation studies on $\delta$ in Table~\ref{tab:ablation_aav} for AAV and Table~\ref{tab:ablation_gfp} for GFP. The performance is robust to the value of $\delta$ and the best performance is obtained when $\delta=0.05$ for AAV and $\delta=0.2$ for GFP. We observed that as $\delta$ increases, fitness tends to decrease while diversity increases, indicating a tradeoff. Therefore, the choice of $\delta$ should be based on the specific priorities of the task.

\begin{table}[h!]
  \centering
  \begin{minipage}[t]{0.48\linewidth}\centering
    \begin{tabular}{ccccc}
      \toprule
      $\delta$ & Fitness $\uparrow$ & Diversity & $d_\text{init}$ & $d_\text{high}$ \\
      \midrule
      $0.05$ & \textbf{0.67} (0.1) & 5.4 (0.5) & 7.0 (0.0) & 4.0 (0.0) \\
      $0.1$ & 0.66 (0.0) & 6.0 (1.2) & 7.0 (0.0) & \textbf{2.0} (0.0) \\
      $0.2$ & 0.63 (0.0) & 8.2 (2.6) & 8.4 (0.5) & 7.0 (0.7) \\
      \bottomrule
    \end{tabular}
    \caption{Ablation studies on $\delta$ in AAV \textit{hard} task.}
    \label{tab:ablation_aav}
  \end{minipage}
  \hfill 
  \begin{minipage}[t]{0.48\linewidth}\centering
    \begin{tabular}{ccccc}
      \toprule
      $\delta$ & Fitness $\uparrow$ & Diversity & $d_\text{init}$ & $d_\text{high}$ \\
      \midrule
      $0.2$ & \textbf{0.88} (0.0) & 4.2 (0.4) & 7.0 (0.0) & 3.0 (0.0)\\
      $0.3$ & 0.85 (0.0) & 4.8 (0.5) & 7.0 (0.0) & \textbf{2.0} (0.0) \\
      $0.5$ & 0.83 (0.0) & 5.0 (0.7) & 7.0 (0.0) & 3.2 (0.4) \\
      \bottomrule
    \end{tabular}
    \caption{Ablation studies on $\delta$ in GFP \textit{hard} task.}
    \label{tab:ablation_gfp}
  \end{minipage}
\end{table}

The parameter $m_\text{step}$ is set to regularize the policy to not take steps that lead to a high number of mutations from the sequence of the current state. The recommended value is 3. 
The parameter $m_\text{decode}$ is set based on the VED accuracy and the sequence length $L$ for the target protein family. This value also affects the maximum number of mutations allowed at each timestep. The parameter $m_\text{total}$ is set based on the protein design specifications for desired optimization sequences. The parameter $T_\text{ep}$ defines the length of the episode. The recommended range of $T_\text{ep}$ is $[3, 8]$ given that we are using a sparse reward modeling in which the reward is given just at the end of the episode. For settings using an in silico predictor in which reward can be calculated at every timestep and a dense reward modeling, this value can be increased allowing the policy to take more steps through the fitness landscape at each episode.

\section{Double-loop Optimization}
\label{app:double}

\looseness=-1 We show that LatProtRL achieves high performance using a black-box oracle in Table~\ref{tab:main} and an in silico predictor in Section~\ref{sec:predictor}. For a setting in which the number of rounds is very limited, \textit{e.g.} 5 rounds, our method adds an option of using an in silico predictor to update the policy in between rounds of black-box evaluation. We investigate a double-loop setting (Algorithm~\ref{alg:double}) where we (i) train the RL policy using the predictor evaluation as a reward in an inner loop, (ii) train the predictor based on oracle evaluation of variants proposed in the final round of the inner loop. We test the double-loop methodology with a limited number of rounds $E=5$ and setting $N_\text{oracle calls}$ to 256. As shown in Table~\ref{tab:single_double}, the double-loop method is effective and outperforms both using only the oracle and only the predictor for 5 rounds.

\begin{algorithm}
\caption{Double-loop optimization}
\label{alg:double}
     \begin{algorithmic}[1]
     \Function{Inner-Loop}{$q, E_\text{inner}$}
        \State Run Alg.~\ref{alg:main} for $E_\text{inner}$ rounds with $q$ (see line 25 of Alg.~\ref{alg:main})
        \State \Return trajectories $\mathcal{T}'$ (see line 29 of Alg.~\ref{alg:main})
     \EndFunction
     \State Predictor $g_\phi$, oracle $g$
     \For{\textbf{5} rounds}
        \State $\mathcal{T}_\text{oracle} \gets $ \Call{Inner-Loop}{$g, 1$}
        \State Train $g_\phi$ with $\mathcal{T}_\text{oracle}$
        \State \Call{Inner-Loop}{$g_\phi, 2$}
     \EndFor 
     \State \Call{Inner-Loop}{$g_\phi, 10$}
     \end{algorithmic}
\end{algorithm}

\begin{table}[t!]
    \centering
    \begin{tabular}{llccc|ccc}
    \toprule
         & & \multicolumn{3}{c}{AAV \textit{medium} task} & \multicolumn{3}{c}{AAV \textit{hard} task} \\
         \cmidrule(lr){3-5} \cmidrule(lr){6-8}
         Method & & Fitness $\uparrow$ & Diversity & $d_\text{init}$ & Fitness $\uparrow$ & Diversity & $d_\text{init}$ \\
     \toprule \multirow{3}{*}{Ours}& \textbf{Double-loop $E=5$} & \textbf{0.70} (0.0) & 4.4 (0.5) & 5.6 & \textbf{0.65} (0.0) & 6.3 (1.2) & 7.3 \\
        & Oracle $E=5$ & 0.60 (0.0) & 8.0 (0.7) & 7.4 & 0.54 (0.0) & 11 (1.1) & 9.0 \\
        & Predictor-only & 0.57 (0.0) & 3.0 (0.0) & 5.0 & 0.57 (0.0) & 3.0 (0.0) & 7.0 \\
    \midrule
        \multirow{2}{*}{AdaLead} & Oracle $E=5$ & 0.66 (0.2) & 3.2 (0.4) & 9.2 & 0.63 (0.0) & 3.0 (0.0) & 10 \\
        & Predictor-only & 0.43 (0.0) & 3.8 (0.4) & 2.0 & 0.44 (0.0) & 2.9 (0.9) & 2.0 \\
    \toprule
        & & \multicolumn{3}{c}{GFP \textit{medium} task} & \multicolumn{3}{c}{GFP \textit{hard} task} \\
        \cmidrule(lr){3-5} \cmidrule(lr){6-8}
        & Method & Fitness $\uparrow$ & Diversity & $d_\text{init}$ & Fitness $\uparrow$ & Diversity & $d_\text{init}$ \\
    \midrule
        \multirow{3}{*}{Ours} & \textbf{Double-loop $E=5$} & \textbf{0.91} (0.0) & 4.2 (0.4) & 5.8 & \textbf{0.81} (0.0) & 4.8 (0.8) & 7.2\\
         & Oracle $E=5$ & 0.90 (0.0) & 4.8 (0.8) & 7.2 & 0.77 (0.1) & 6.2 (1.3) & 8.4 \\
        & Predictor-only & 0.81 (0.0) & 3.0 (0.0) & 5.0 & 0.75 (0.0) & 3.0 (0.0) & 7.0 \\
        \midrule 
        \multirow{2}{*}{AdaLead} & Oracle $E=5$ & 0.80 (0.0) & 3.6 (1.1) & 15 & 0.50 (0.0) & 3.2 (1.3) & 17 \\
        & Predictor-only & 0.59 (0.0) & 5.5 (0.6) & 2.0 & 0.39 (0.0) & 3.5 (0.6) & 2.0 \\
    \bottomrule
    \end{tabular}
    \caption{\textbf{Single vs. double-loop optimization results}. The standard deviation of 5 runs with different seed is indicated in parentheses.}
    \label{tab:single_double}
\end{table}

\section{Variant Encoder-Decoder}

We provide more information regarding the VED architecture and training. We used the pre-trained ESM-2~\cite{lin2022language} model with 650M parameters for both the encoder and decoder. The ESM-2 650M model consists of a token embedding layer and 33 transformer layers. Each transformer layer consists of multi-head attention followed by a layer normalization layer and two fully connected layers with GeLU \citep{hendrycks2016gaussian} non-linear activation. The language model head comprises a linear layer followed by GeLU activation, and the linear layer uses the weight of the embedding layer in the encoder. The number of focus heads is set to 20. Since we provide a reference representation during sequence recovery and mutants in the GFP dataset have a maximum of 15 mutations, we add a constrained decoding objective during inference time to restrict the number of mutations in the input sequence. We train a separate VED for each of the four tasks evaluated in this work. We held 5\% of the data as a test dataset before augmentation. The training dataset is augmented by 4 times using random mutations, where the expected number of mutations is set to 3. The model is trained on the training set of each dataset for 32 epochs using the Adam optimizer \citep{kingma2014adam} for GFP and 64 epochs for AAV. The initial learning rate of Adam is set to 1e-3, with weight decay set to 1e-5. Table \ref{tab:seqs} shows the performance of a sequence decoder.

\begin{table}[h]
    \centering
    \begin{tabular}{l|c|c}
    \toprule
    \multirow{2}{*}{Dataset} & \multicolumn{2}{c}{Top-1 Accuracy} \\
    & Mutated positions & Non-mutated positions \\
    \midrule
    GFP \textit{medium} & 0.40 & 0.95 \\
    GFP \textit{hard} & 0.43 & 0.94 \\
    AAV \textit{medium} & 0.62 & 0.83 \\
    AAV \textit{hard} & 0.57 & 0.78 \\
    \bottomrule
    \end{tabular}
    \caption{Decoding accuracy of VED on test datasets. Mutated positions are with respect to the reference sequence.}
    \label{tab:seqs}
\end{table}

As shown in Table~\ref{tab:seqs}, the decoder architecture is still an open question, and improvement in the decoding accuracy is desired. With the use of the proposed constrained decoding strategy, the decoding step can be thought of as mutations proposed by the policy with a rather inherent degree of exploration.

\section{Oracle and Training Details}

\paragraph{Oracle}
The oracle proposed in \citet{kirjner2023optimizing} is based on a convolutional neural network (CNN) architecture. This architecture uses 1-dimensional convolutional layers with 256 channels taking as input a one-hot encoding of the protein sequence. This layer is followed by a max-pooling and a dense layer to output a single value representing the predicted fitness. The oracle prediction shows a Spearman correlation of 0.89 for GFP dataset. This oracle is used for evaluation in all benchmarks. For the experiments using the fitness predictor, we use the implementation and results of \citet{kirjner2023optimizing}. 

\paragraph{Policy}
The RL policy is trained using PPO \citep{schulman2017proximal}, an on-policy RL algorithm. For the experiments,  we use the implementation provided by \href{https://stable-baselines.readthedocs.io/en/master/}{Stable Baselines} \citep{stable-baselines}. The default hyperparameters of the PPO class are used to train the RL policies. In our formulation, the action space is continuous.

\paragraph{Optimization using a fitness predictor}
We ran a total of 15,000 timesteps for GFP \textit{hard} and 20,000 timesteps for the other three tasks and reported the result of the last evaluation round. Other hyperparameters regarding policy training are similar to the experiments using the oracle except for the fact that the reward is calculated for every timestep when using the predictor.

\section{Policy avoiding Local Optima}
\label{app:localoptima}

\begin{figure}[h!]
    \centering
    \includegraphics[width=0.6\linewidth]{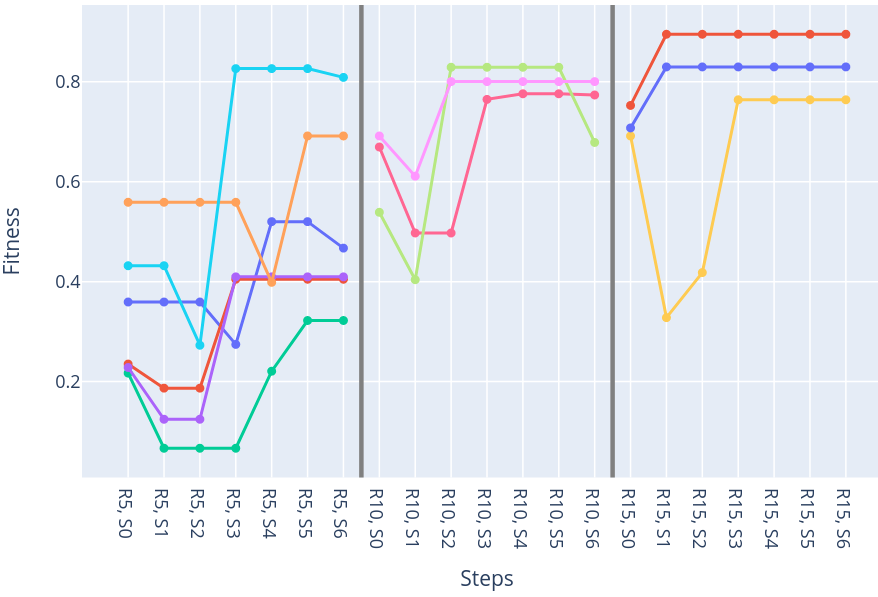}
    \caption{\textbf{Sample trajectories from the RL policy during GFP \textit{hard} optimization} for rounds 5, 10, and 15.}
    \label{fig:traj}
\end{figure}

In Figure~\ref{fig:traj}, we show selected trajectories from the RL policy during optimization for the GFP \textit{hard} task. To propose sequences that achieve higher fitness when compared to the initial sequence, in specific cases the policy takes actions that lead to sequences with lower predicted fitness during an episode. For example, in round 5, 37 out of 256 trajectories increased fitness, with 12 out of those 37 trajectories having decreasing steps. We show 5 sample trajectories with decreasing steps in Figure~\ref{fig:traj}. In round 10, 28 out of 256 trajectories increased fitness, with 7 out of those 28 trajectories having steps decreasing fitness values. We show 3 sample trajectories with decreasing steps in Figure~\ref{fig:traj}. In the final round, since the fitness is already high and the set of starting sequences is close to the experimental distribution of high-functional variants, only one trajectory had decreasing steps. This trajectory is shown in yellow in Figure~\ref{fig:traj}. We show two additional trajectories (red and blue) that increase fitness at round 15 without decreasing steps showing the policy optimization behavior.

\end{document}